\documentclass[journal,10pt]{IEEEtran}
\usepackage{geometry}
\usepackage{graphicx}
\usepackage[cmex10]{amsmath}
\usepackage{subfigure}
\usepackage{stfloats}
\usepackage{epstopdf}
\usepackage{cite}
\usepackage{algorithmic}
\usepackage[ruled,linesnumbered]{algorithm2e}  
\usepackage{booktabs}
\usepackage{multirow}
\usepackage[table,xcdraw]{xcolor}
\usepackage{makecell, multirow, diagbox}

\usepackage{amsmath}
\usepackage{amssymb}
\usepackage{color}
\usepackage{xcolor}
\usepackage{xpatch}
\usepackage{bm}

\makeatletter
\ExplSyntaxOn
\cs_new:Npn \bibColoredItems #1#2
  {
    \clist_map_inline:nn {#2} { \cs_new:cpn {bib@colored@##1} {#1} } 
  }
\ExplSyntaxOff

\newcommand\bib@setcolor[1]{%
  \ifcsname bib@colored@#1\endcsname
    \expanded{\noexpand\color{\csname bib@colored@#1\endcsname}}%
  \else
    \normalcolor
  \fi
}

\xpatchcmd\@bibitem
  {\item}
  {\bib@setcolor{#1}\item}
  {}{\fail}

\xpatchcmd\@lbibitem
  {\item}
  {\bib@setcolor{#2}\item}
  {}{\fail}
\makeatother

\title{Cross-Fusion Rule for Personalized\\ Federated Learning}
\author{Wangzhuo~Yang, Bo~Chen, Yijun~Shen, Jiong~Liu, Li~Yu
\thanks{W. Z. Yang, B. Chen, Y. J. Shen, J. Liu and L. Yu are with the Department of Automation, Zhejiang University of Technology, Hangzhou 310023, China, and also with the Institute of Cyberspace Security, Zhejiang University of Technology, Hangzhou 310023, China (Correspondence email:bchen@aliyun.com).}}

\begin{document}

\markboth{xxxx-xxxx-xxxx}
{Shell \MakeLowercase{\textit{et al.}}: Bare Demo of IEEEtran.cls for Journals}

\maketitle
\begin{abstract}
Data scarcity and heterogeneity pose significant performance challenges for personalized federated learning, and these challenges are mainly reflected in overfitting and low precision in existing methods.
To overcome these challenges, a multi-layer multi-fusion strategy framework is proposed in this paper, i.e., the server adopts the network layer parameters of each client upload model as the basic unit of fusion for information-sharing calculation.
Then, a new fusion strategy combining personalized and generic is purposefully proposed, and the network layer number fusion threshold of each fusion strategy is designed according to the network layer function.
Under this mechanism, the $L_2$-Norm negative exponential similarity metric is employed to calculate the fusion weights of the corresponding feature extraction layer parameters for each client, thus improving the efficiency of heterogeneous data personalized collaboration.
Meanwhile, the federated global optimal model approximation fusion strategy is adopted in the network full-connect layer, and this generic fusion strategy alleviates the overfitting introduced by forceful personalized.
Finally, the experimental results show that the proposed method is superior to the state-of-the-art methods.
\end{abstract}

\begin{IEEEkeywords}
personalized federated learning, layer-based, cross-fusion rule, multi-layer multi-fusion strategy, heterogeneous data.
\end{IEEEkeywords}

\section{Introduction}
With the increasing emphasis on data privacy, federated learning without sharing private data has attracted more scholarly attention \cite{a1, kbs1}.
Because of the non-shared data feature that protects user privacy, federated learning is commonly utilized in the financial, medical, industrial, and energy domains \cite{a2, kbs2, a4, a28, a29}.
While there has been a significant amount of work focusing on the optimization aspects of federated learning, model overfitting and data heterogeneity remain key challenges to be faced in federated learning \cite{a5, a6, a25, a26, kbs3}.
The heterogeneous data with non-independent and identically distributed makes the model obtained by each client one-sided, which cannot represent the integrity of the sample space.
In this context, the argument that data heterogeneity exacerbates the overfitting of the model is held in \cite{a7}. 
Therefore, how to design reliable algorithms for the client-side data heterogeneity problem is the essence of studying federated learning.

To address this problem, a global model re-parameterization idea was suggested in \cite{fedprox}, which uses the fusion model as a constraint target for the client, causing the client model to be trained to approximate that target.
However, this method does not consider the collaboration type, resulting in similar clients not effectively using model information of each other for performance improvement.
In this case, some personalized federated learning (PFL) methods have been proposed by \cite{a24, a8, a9, a10, a12, a13}, which are dedicated to optimising the collaboration strategy between client models.
Specifically, a similar strategy to \cite{fedprox} is used in \cite{a12}, with the improved aspect of changing the global model as the penalty target to a private model for each client.
Further, the pFedMe proposed in \cite{a13} uses the Moreau envelopes as a constraint term to enable personalized deployment of the client model, where a bi-level strategy was adopted to calculate the server model of conditional convergence.
Based on the strategy of adding additional terms, the improved PFL method compensates for the weakness of personalized data processing in \cite{a5}, which was caused by pursuing the global optimum through the weighted average client model.
Unfortunately, the lack of processing strategies for data heterogeneity in server fusion has led to a general problem of low precision in these methods.
Accordingly, it is necessary to design a fine-grained fusion policy to calculate the personalized weights of each client.
Under this case, a method named personalized federated few-shot learning was developed \cite{a14}, and the core idea of this method is to construct a client personalization feature space, where the feature similarity is calculated as a metric to determine the fusion weight of each client.
Furthermore, the fusion model is utilized as the penalty term of the corresponding client to improve the personalization capability.
In \cite{a15}, a novel client-personalized weight calculation strategy is developed on the server side, i.e., the weighting factors are designed as negative exponential mapping distances between each other's models. 
Then, the personalized model of each client is obtained by the weighting calculation, which will be used as a penalty factor for the optimization objective to improve the model performance.
Similarly, a weighting strategy with a different structure is also proposed in \cite{pfedfomo}, which uses the first-order extreme value points of the model loss function to approximate the optimal weights.

However, the sophisticated personalized fusion rules set in these PFL methods lead to the risk of overfitting while solving the data heterogeneity problem.
To mitigate the overfitting phenomenon in federated learning, a method of Gaussian processes-based PFL was proposed in \cite{a16}, which has a better representation of the model due to the nature of Bayesian.
Analogously, a PFL method based on Bayesian neural networks was proposed \cite{a6}, where the penalty term is denoted as the KL distance between the hypothetical distribution and the posterior distribution of the model parameters.
Notice that both the kernel function selection in the Gaussian process and the parameter hypothetical distribution in the Bayesian neural network depends on a large amount of data support, which makes these two methods unsuitable for scenarios with a small amount of heterogeneous data.
Therefore, the overfitting problem of heterogeneous data with few samples is still a problem that each client needs to focus on.

To meet the data heterogeneity requirements of different clients, global efficiency and local model personality issues are considered in \cite{a17}.
This inspires us to design the global model as a generic term to solve the overfitting problem and the personalized model as a penalty term to meet the personalized needs of heterogeneous data, respectively.
Meanwhile, an idea of fusion based on the neural network layers was suggested in \cite{a18}.
And then, a layer-based federated learning method was developed in \cite{a19}, which requires a portion of the raw data for the server to train the fused weights. 
Although this method defeats the original purpose of federated learning data preservation, it further reminds us to integrate the generic and personality terms into one model, where the network layer is treated as the most basic fusion unit.
Moreover, a deep neural network is considered by \cite{a27} to be divided into two parts, shallow and deep, and it is noted that they are generic and specialized, respectively, further inspiring the design of a functional layer-based fusion strategy in this paper.

Motivated by the above analysis, we shall study the data heterogeneity and overfitting problem for federated learning systems.  
Unlike traditional PFL, which is dedicated to personalized item design, instead, both personalized and generic items are focused on in this work.
The main contributions of this paper are as follows.
\begin{itemize}
\item [1)] For the PFL systems, a novel layer-based personalized federated fusion rule, which is different from the pseudo-federated structure in \cite{a19}, is proposed by combining different fusion policies employed at different network layers in each communication epoch. 
Then, a personalized fusion framework for multi-layer multi-fusion strategies is presented in this paper. 
Subsequently, a rule to determine the fusion threshold of the number of network layers for each fusion strategy is designed based on the network layer function.
\item [2)] Based on a multi-layer multi-fusion framework, a strategy for cross-fusion of personalized and generic is implemented in this paper.
According to the negative exponential distance mapping of $L_2$-Norm similarity metric, the rule for calculating the fusion weights between clients is improved to achieve personalized fusion of the feature extraction layer.
On the other side, a generic term of federated global optimal model approximation fusion strategy for the network full-connect layer is used to alleviate the overfitting phenomenon of the client.
It should be stressed that the personalized and generic terms in this paper refer to the processing rules of different layers in the model fusion, which through their respective properties enhance the model performance.
\item [3)] The layer function-based fusion threshold rule is applied to the multi-layer multi-fusion strategy framework to improve personalized federated learning performance.
Then, the extensive experiments on three benchmark datasets show that the proposed personalized federated learning based on the cross-fusion rule (pFedCFR) outperforms state-of-the-art (SOTA) PFL methods \cite{a12, a13, a15} and generic federated learning strategy \cite{a5, fedprox}.
\end{itemize}

\section{Problem Formulation}
Consider a federated learning system with $N$ clients described by the following structure:
\begin{equation}\begin{aligned}\label{e1}
\left\{
\begin{array}{lc}
\text{Client}: c_1,...c_n,...c_N\\
\text{Parameter}:\bm{\nu_1},...\bm{\nu_n},...\bm{\nu_N}\\
\text{Dataset}: \mathcal{D}_1,...\mathcal{D}_n,...\mathcal{D}_N
\end{array}
\right.
\end{aligned}\end{equation}
where the structure of network model $M$ is the same for all clients, and thus the size of the corresponding model parameter $\bm{\nu}_n$ is the same for each $c_n$, $\mathcal{D}_n$ is a private training dataset for each $c_n$ that is non-independently and identically distributed.
For each $c_n$, the best performance of $M(\bm{\nu}_n)$ on the $\mathcal{D}_n$ is illustrated by $\bm{\nu}_n^*$.

Specifically, each client $c_n$ individually represents the loss of model parameter $\bm{\nu}_n^*$ in the training dataset $\mathcal{D}_n$ through a cost function $\{\mathcal{F}_n(\bm{\nu}_n):R^d \to R, \bm{\nu}_n \in R^d\}$.
Thus, through the collaboration of each client, the goal of personalized federation learning is then defined as follows:

\begin{equation}\begin{aligned}\label{e2}
\min_{V} G(V):= \sum_{n=1}^N \mathcal{F}_n(\bm{\nu}_n) + \mathcal{P}(V)
\end{aligned}\end{equation}
where $V$ denotes the parameter collection of each client, i.e. $V=[\bm{\nu_1},...\bm{\nu_n},...\bm{\nu_N}]$, $G(V)$ is the global optimization object, and $\mathcal{P}(V)$ is the penalty term to each client.
The loss term $\mathcal{F}_n(\bm{\nu}_n)$ of each $c_n$ in (\ref{e2}) is calculated by its personalized training dataset.
Meanwhile, the parameter $\bm{\nu}_n$ is updated and transmitted to the server. 

Based on the parameter $\bm{\nu}_n$, the basic unit of collaboration information is given by
\begin{equation}\begin{aligned}\label{e3}
\bm{\nu}_n=[\bm{\nu}_{n,0},\cdots,\bm{\nu}_{n,l},\cdots,\bm{\nu}_{n,L}]^T
\end{aligned}\end{equation}
where $\bm{\nu}_{n,l}$ denotes the model parameters of $l$th layer, $L$ is the depth of the model, and the collaborative way will be designed in Section III.

Subsequently, with the consideration of data heterogeneity and model overfitting, the cross-fusion strategy structure of each client is given by
\begin{equation}\begin{aligned}\label{e4}
\text{RULE} : fusion_{p} \Rightarrow fusion_{g}
\end{aligned}\end{equation}
where symbol $\Rightarrow$ indicates that the cross-fusion rule consists of two serial fusion rules, and the personalized fusion rule $fusion_p$ and general fusion rule $fusion_{g}$ will be designed in Section III.
Consequently, the issues to be addressed in this paper are described as follows.
\begin{itemize}
\item [1)] The first aim is to design a layer-based personalized federated fusion structure for (\ref{e3}) such that collaboration information $\bm{\nu}_{n,l}$ is more granular, and the key information interactions independent of each other.
\item [2)] Under the cross-fusion strategy (\ref{e4}), the second aim is to design the personalized fusion rule $fusion_{p}$ and general fusion rule $fusion_{p}$, such that the raw data feature extraction layers $[\bm{\nu}_{n,0}:\bm{\nu}_{n,l}],l<L$ have strong personalization capability, and the remaining layers have generalization capability.
\end{itemize}

\textbf{Remark 1:} 
It is concluded from (\ref{e2}) that the penalty term $\mathcal{P}(V)$ directly affects the optimization objective of the proposed cross-fusion rule.
Through the collaboration between model parameters in $V$, penalty terms $\mathcal{P}(V)$ are computed, improving the performance of the $c_n$ under heterogeneous dataset $D_n$.
Notice that the core of PFL is the collaborative strategy among the clients, while the penalty term aims at deep optimization of $\bm{\nu}_n$ in \cite{a8, fedprox, a13, a15}, which implies us to design fusion rule with considering data heterogeneity and model overfitting in this paper for each client.
Moreover, when designing the fusion rule in this paper, each client's layer parameter is proposed to be viewed as the basic unit.
This also inspires us to focus on the influence of using different fusion strategies at the same layer.
 
\textbf{Remark 2:} 
It is known from (\ref{e4}) that there are 2 serial fusion rules in the server.
Combined with the layer-based fusion structure in (\ref{e3}), fusion rule $fusion_{p}$ and $fusion_{g}$ are designed to handle the information collaboration of different layer parameters between clients, respectively.
It should be noted that the rule $fusion_{p}$ is dedicated to the personalization study of heterogeneous data, while $fusion_{g}$ aims to solve the overfitting problem.
In this case, the $\bm{\nu}_{n}$ obtained by fusion is also present in the penalty term with the same shape.

\textbf{Notations:} Since the server needs to calculate the weights of each local model in federation learning, fusion is considered more appropriate than aggregation in this paper.
The superscript “T” represents the transpose, while $Diag()$ denotes extracting the elements on the diagonal of the matrix and forming the column vector.
The symbol “$\to$” indicates a point-to-point connection.
$\lambda$, $\mu$ and $\alpha_t$ are hyperparameter greater than 0.

\section{Proposed method}  
\begin{figure*}[!htb]
	\setlength{\abovecaptionskip}{0cm}
	\centering
	\includegraphics[width=16.8cm,height=9.6cm]{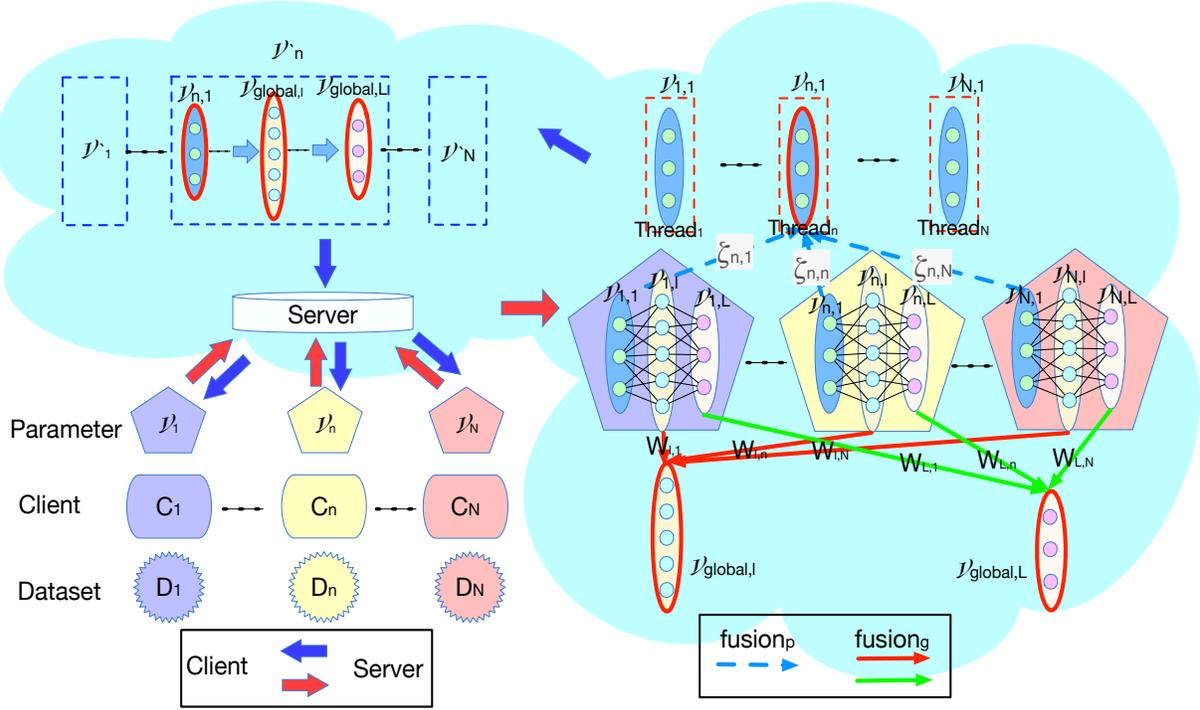}
	\caption{Mechanistic framework of pFedCFR. Each round of iteration proceeds as follows.
	1). \textbf{Upload}: Each client uploads the trained model to the server, and the collaboration of client model parameters is implemented in the server with layers as the basic unit. 
	2). \textbf{Fusion}: There are two types of fusion: $fusion_{p}$, represented by the blue dashed arrow, and $fusion_{g}$, represented by the red and green solid arrows.
	Each client in $fusion_{p}$ obtains its personalized network parameters by calculating weighting factors, while all clients in $fusion_{g}$ share a layer of network parameters.
	3). \textbf{Allocation}: Restructure the network parameters of the personalization layer $\bm{\nu}_{n,1}$ and the generic layer $\bm{\nu}_{global,l}$, and distribute the restructured parameters $\bm{\nu}'_{n}$ to the corresponding client $c_n$.}
	\label{fig1}
\end{figure*}

In this section, a layer-based client collaboration idea is improved to enhance refinement processing capability, based on which a personalized fusion framework with multi-layer multi-fusion strategies is designed.
Then, a threshold calculation rule involving the number of network layers under each fusion strategy is designed.
Furthermore, a cross-fusion rule will be developed to tackle the problem of data heterogeneity and overfitting in the PFL background.
For more visualization, the overall framework of the proposed pFedCFR is shown in Fig.~{\ref{fig1}}. 

\subsection{Layer-Based Structure}
Conventional methods achieve collaboration between clients by weighted fusion of overall models, but these methods ignore the specificity of network layer functionality and its varying roles under different models.
To solve this problem, a layer-based fusion structure is developed in this paper, using neural network layers as the basis fusion unit.
Let $\bm{\omega}_l=[{\omega}_{1,l}, {\omega}_{2,l},\cdots,{\omega}_{N,l}]^T$ and $W_n =[\bm{\omega}_1,\bm{\omega}_2,\cdots,\bm{\omega}_L] $, where $\omega_{n,l}$ denotes the weight of $l$th layer in $c_n$, and its specific value is given by the fusion rule, where the subscript $n$ in $W_n$ indicates that the current computational sequence is $c_n$.
Then, combining with (\ref{e3}), the fusion result for each epoch of the $c_n$ is expressed as
\begin{equation}
\begin{aligned}\label{e5}
\bm{\nu}'_{n} &= Diag(V \cdot W_n)\\ &=\!Diag\!\!
\left(\!
\left[\!\!\!
\begin{array}{cccc}
\bm{\nu}_{1,1}& \!\!\!\!\bm{\nu}_{2,1}& \!\!\!\!\cdots&  \!\!\!\!\bm{\nu}_{N,1}\\
\bm{\nu}_{1,2}& \!\!\!\!\bm{\nu}_{2,2}& \!\!\!\!\cdots&  \!\!\!\!\bm{\nu}_{N,2}\\
\vdots & \vdots & \ddots & \vdots \\ 
\bm{\nu}_{1,L}& \!\!\!\!\bm{\nu}_{2,L}& \!\!\!\!\cdots&  \!\!\!\!\bm{\nu}_{N,L}
\end{array}
\!\!\!\right]
\!\!\!
\left[\!\!\!
\begin{array}{cccc}
\omega_{1,1} & \!\!\!\!\omega_{1,2} & \!\!\!\!\cdots & \!\!\!\!\omega_{1, L}\\
\omega_{2,1} & \!\!\!\!\omega_{2,2} & \!\!\!\!\cdots & \!\!\!\!\omega_{2, L}\\
\vdots & \vdots & \ddots & \vdots \\
\omega_{N,1} & \!\!\!\!\omega_{N,2} & \!\!\!\!\cdots & \!\!\!\!\omega_{N, L}
\end{array}
\!\!\!\right]
\!\right)\\
&=\left[\bm{\nu}'_{n,1},\bm{\nu}'_{n,2},\cdots,\bm{\nu}'_{n,L}\right]^T
\end{aligned}\end{equation}
where $\bm{\nu}'_{n,l}$ denotes the updated layer parameters, whose value is $\omega_{1,l} \cdot \bm{\nu}_{1,l}  +\cdots+ \omega_{N,l} \cdot \bm{\nu}_{N,l}$.

In the existing PFL algorithms \cite{fedprox, a15}, the weight $\omega^G_{n,l}$ contained in $\bm{\nu}_n$ takes the same value, while each $\omega_{n,l}$ in (\ref{e5}) is more flexible and diverse.
Assuming that the fusion strategy is denoted by $fusionRule$, the weights of the two approaches can be expressed as
\begin{equation}\begin{aligned}\label{e6}
\left\{
\begin{array}{l}
\omega^G_{n,1}=\cdots=\omega^G_{n,L}\triangleq fusionRule(\bm{\nu}_n;\bm{\nu}_1,\bm{\nu}_2,\cdots,\bm{\nu}_N)\\
\omega_{n,l}\triangleq  fusionRule(\bm{\nu}_{n,l};\bm{\nu}_{1,l},\bm{\nu}_{2,l},\cdots,\bm{\nu}_{N,l})
\end{array}
\right.
\end{aligned}\end{equation}
Notice that $\omega_{n,l}$ is obtained by only computing the $l$th layer network parameters, which implies that multiple fusion rules can exist for a federated learning algorithm based on layers as fusion units.
Following this idea, a multi-layer multi-fusion strategy structure is developed to be given by
\begin{equation}\begin{aligned}\label{e7}
\left\{
\begin{array}{lc}
fusionRule_1(\bm{\nu}_{1,1},\bm{\nu}_{2,1},\cdots,\bm{\nu}_{N,1})\\
fusionRule_2(\bm{\nu}_{1,2},\bm{\nu}_{2,2},\cdots,\bm{\nu}_{N,2})\\
\;\;\;\;\;\;\;\;\;\;\;\;\;\;\;\;\;\;\;\;\;\;\;\;\vdots\\
fusionRule_L(\bm{\nu}_{1,L},\bm{\nu}_{2,L},\cdots,\bm{\nu}_{N,L})\\
\end{array}
\right.
\end{aligned}\end{equation}
where $fusionRule_l$ is the adopted fusion rule of $l$th layer.

\textbf{Remark 3:} 
It can be seen from (\ref{e6}) that $\omega_{n,l}$ is co-determined by $[\bm{\nu}_{1,l},\bm{\nu}_{2,l},\cdots,\bm{\nu}_{N,l}]$, which is more refined and targeted than $\omega^G_{n,l}$ when considering that each layer of the network has a different impact on the model.
Since the optimization target differs under different fusion strategies, each fusion rules $fusionRule_l$ in (\ref{e7}) can be the same or different.
Meanwhile, considering that the functions of each network layer in the model vary, a method for determining the fusion threshold based on the network layer function was developed.

\subsection{Function-Based Fusion Threshold Rule}
Given the variability of functions among network layers in the deep learning framework and the different focus of each fusion strategy, a rule to determine the threshold of the network layers under each fusion strategy based on the layer functions is developed in this subsection.
According to the multi-layer multi-fusion strategies structure of (\ref{e7}), it is assumed that the network layers $1$ to $l_1$, $l_1+1$ to $l_2$, $\cdots$, $l_n$ to $L$ each have the same function, where $1<l_1<l_2\cdots<l_n<L$.
Then, the function-based fusion threshold rule with multi-fusion strategies is proposed to be
\begin{equation}\begin{aligned}\label{eq1}
\left\{\!\!\!\!
\begin{array}{lc}
fusionRule_1=\cdots=fusionRule_{l_1},r_1=[1,l_1]\\
fusionRule_{l_1+1}=\cdots=fusionRule_{l_2},r_2=[l_1+1,l_2]\\
\;\;\;\;\;\;\;\;\;\;\;\;\;\;\;\;\;\;\;\;\;\;\;\;\;\;\;\;\;\;\;\;\;\;\;\;\;\;\;\;\;\vdots\\
fusionRule_{l_n+1}=\cdots=fusionRule_{L},r_{n+1}=[l_n+1,L]\\
\end{array}
\right.
\end{aligned}\end{equation}
where $r_i(0<i<n+1)$ is the threshold of each fusion strategy. 
With this mechanism, fusion strategies that match the characteristics of the network layer can be targeted to improve the performance of the federated learning model.

\subsection{Cross-Fusion Rule}
Based on the framework of multi-layer multi-fusion strategy, the combination of two fusion strategies, forceful personalized and generic, will be presented as follow.
Concretely, to solve the problems of data heterogeneity and model overfitting, a pFedCFR structure is proposed in this section, which improves the generic strategy in \cite{fedprox} and the personalized strategy in \cite{a15}, respectively.
At the same time, according to the strategy in (\ref{e4}), $fusion_{p}$ is designed as a personalized fusion rule for the raw data feature extraction layer, while $fusion_{g}$ is a fusion rule for the generic full-connect decision layer.
Then, the result of $fusion_{p}$ is set as the input of $fusion_{g}$ in forward propagation.
In this case, the optimization problem in (\ref{e2}) is rewritten as
\begin{equation}\begin{aligned}\label{e8}
\arg \min_{\bm{\nu}_n} \mathcal{L}(\bm{\nu}_n):=  \mathcal{F}_n(\bm{\nu}_n) + \sum_{l=1}^L\mathcal{P}_l(\bm{\nu}_{n,l})
\end{aligned}\end{equation}
The specific details of the two fusion rules are shown below.

\subsubsection{Personalized Fusion Rule}
As shown in Fig.~{\ref{fig1}}, a thread is opened on the server for each client, which can read all shared $[\bm{\nu}^t_{1,l},\bm{\nu}^t_{2,l},\cdots,\bm{\nu}^t_{N,l}]$.
Then, according to the message passing mechanism in \cite{a15}, the layer-based personalized fusion rule $fusion_p$ in (\ref{e4}) is given by
\begin{equation}\begin{aligned}\label{e9}
\bm{\nu}^t_{n,l} 
& =\omega_{n, 1, l} \bm{\nu}^{t-1}_{1,l}\cdots+\omega_{n, m, l} \bm{\nu}^{t-1}_{n,l}\cdots+\omega_{n, N, l} \bm{\nu}^{t-1}_{N,l}\\
& =\left(1-\alpha_{t} \sum_{m \neq n}^{N} A^{\prime}\left(\left\|\bm{\nu}^{t-1}_{n,l} - \bm{\nu}^{t-1}_{m,l}\right\|^{2}\right)\right) \cdot \bm{\nu}^{t-1}_{n,l} \\ 
& +\alpha_{t} \sum_{m \neq n}^{N} A^{\prime}\left(\left\|\bm{\nu}^{t-1}_{n,l} - \bm{\nu}^{t-1}_{m,l}\right\|^{2}\right)\cdot \bm{\nu}^{t-1}_{m,l}
\end{aligned}\end{equation}
where $\omega_{n, 1, l},\cdots,\omega_{n, N, l}$ are collaboration weights of each other in l$th$ layer, $A(x)=1-e^{{-x}/\sigma}$, $A^{\prime}$ is the derivative of $A(x)$ and $\sigma$ is a hyperparameter. 

It can be seen from (\ref{e9}) that when the $l$th layer parameter $\bm{\nu}^{t-1}_{n,l}$ in $c_n$ is more similar to $\bm{\nu}^{t-1}_{m,l}$ in $c_m$, the greater the weighted influence between them.
Thus, the collaboration of clients with similar raw feature spaces is enhanced without exposing private data.
Conversely, the fusion weights $\omega_{n, m, l}$ are inversely proportional for those with large layer parameter distances, 
i.e., there exists a large $||\bm{\nu^{t-1}}_{n,l} - \bm{\nu}^{t-1}_{m,l}||^{2}$ such that little collaborative information interaction between $c_n$ and $c_m$.
Therefore, the personalized weight in Fig.~{\ref{fig1}} was improved as $\zeta_{n,m}=\alpha_{t} A^{\prime}\left(\left\|\bm{\nu}^{t-1}_{n,l} - \bm{\nu}^{t-1}_{m,l}\right\|^{2}\right)$, where $n \neq m$.

Following (\ref{e9}), the penalty term $\mathcal{P}_l(\bm{\nu}_{n,l})$ in (\ref{e8}) is proposed to be
\begin{equation}\begin{aligned}\label{e10}
\mathcal{P}_l(\bm{\nu}_{n,l}) = \frac{\lambda}{2\alpha_t} \left\| \bm{\nu}_{n,l} - \bm{\nu}^t_{n,l}  \right\|^{2}
\end{aligned}\end{equation}
Through this term, $\bm{\nu}_{n,l}$ is forced to approximate $\bm{\nu}^t_{n,l}$, thus achieving the personalized requirements of the optimization objective $\mathcal{L}(\bm{\nu}_{n})$.

\subsubsection{Generic Fusion Rule}
What cannot be ignored is the overfitting problem introduced by the above $fusion_p$ in solving the personality problem of heterogeneous data.
To address this problem, a generic fusion rule $fusion_g$ is developed, which dedicates to information collaboration of the generic full-connect layer. 
As the $\bm{\nu}_{global,l}$ showed in Fig.~{\ref{fig1}}, the designed $fusion_g$ differs in structure from $fusion_p$ in that all clients share $\bm{\nu}_{global,l}$ at layer $l$.
According to the fusion idea in \cite{fedprox}, the layer-based generic fusion strategy is proposed to be
\begin{equation}\begin{aligned}\label{e11}
\bm{\nu}_{global,l} = \frac{1}{N} \sum_{n=1}^N \bm{\nu}_{n,l}
\end{aligned}\end{equation}
where $\bm{\nu}_{global,l}$ is obtained by averaging the cumulative sum of all client parameters at the $l$th layer.
Based on this operation, the impact of each client is equivalent, such that the fusion result exhibits good generalizability.

Subsequently, it is obtained from (\ref{e11}) that the penalty term in this layer can be calculated by 
\begin{equation}\begin{aligned}\label{e12}
\mathcal{P}_l(\bm{\nu}_{n,l}) = \frac{\mu}{2} \left\| \bm{\nu}_{n,l} - \bm{\nu}^t_{global,l}  \right\|^{2}
\end{aligned}\end{equation}
Similarly, $\bm{\nu}_{n,l}$ is forced to approximate $\bm{\nu}^t_{global,l}$ in this layer.

Then, following (\ref{e4}), the optimization objective is decomposed to each layer is denoted as
\begin{equation}\begin{aligned}\label{e13}
\mathcal{F}_n(\bm{\nu}_n) &= \left[ f_{n,1}(\bm{\nu}_{n,1}) \to f_{n,2}(\bm{\nu}_{n,2}) \cdots \to f_{n,L}(\bm{\nu}_{n,L})\right]\\
&=\left[ f_{n,l}(\bm{\nu}_{n,l}) \to\right]_{l=1}^{L}
\end{aligned}\end{equation}
where $f_{n,l}$ denotes the calculation of $c_n$ at layer $l$. 

It is implied from (\ref{e10}) (\ref{e12}) and (\ref{e13}) that the loss function $\mathcal{L}(\bm{\nu}_{n})$ in (\ref{e8}) can be expressed as
\begin{equation}\begin{aligned}\label{e14}
\mathcal{L}(\bm{\nu}_n) &=\left[ f_{n,l}(\bm{\nu}_{n,l}) \to\right]_{l=1}^{L} + \sum_{i=1}^r \frac{\lambda}{2\alpha_t} \left\| \bm{\nu}_{n,i} - \bm{\nu}^t_{n,i} \right\|^{2} \\
& + \sum_{i=r+1}^L \frac{\mu}{2} \left\| \bm{\nu}_{n,i} - \bm{\nu}^t_{global,i}  \right\|^{2}
\end{aligned}\end{equation}
where $r$ is the layer number of the adopted $fusion_p$.
Here, the optimization objective of each client in pFedCFR is obtained.
\begin{algorithm}
\caption{pFedCFR}
\textbf{Notation:} N clients, a private dataset is held by each client; hyperparameter $\mu$, $\alpha$, $\lambda$, model depth $L$, personalized network layers' fusion threshold $r$, communication round $T$ and learning rate $\eta$ are preset to be given.  \\
\textbf{client:} Randomly initialize model parameters $\bm{\nu}_{n}$.\\
\For{ $t=1, 2, \cdots, T$ }{
\textbf{client:} \\
Each client optimize $\bm{\nu}^t_{n}$ by minimizing the loss function $\mathcal{L}(\bm{\nu}^t_{n})$ in (\ref{e14}), and then send the obtained $\bm{\nu}^t_{n}$ to the server.\\
\textbf{server:} \\
\For{ $l=1, 2, \cdots, L$}{
If the current operating network layer is less than $r$, then $\bm{\nu}^{t+1}_{n,l}$ is obtained using the personalized fusion rule (\ref{e9}).\\
Otherwise, the $\bm{\nu}^{t+1}_{n,l}$ is calculated through the generalized fusion rule (\ref{e11}).\\
Recombining $\bm{\nu}^{t+1}_{n} \gets [\bm{\nu}^{t+1}_{n,1},\bm{\nu}^{t+1}_{n,2},\cdots,\bm{\nu}^{t+1}_{n,L}]$.
}
Send $\bm{\nu}^{t+1}_{n}$ to the corresponding $c_n$.
}
\textbf{Output:} $[\bm{\nu}^T_{1},\cdots,\bm{\nu}^T_{N}]$
\end{algorithm}

\textbf{Remark 4:} 
To determine the network layers' fusion threshold $r$ in Algorithm 1, the fusion rule of the feature extraction layer and the full-connect layer in the model is focused on in this paper.
Noting that the message passing mechanism in (\ref{e9}) enhances the influence between similar feature layers, the designed personalized fusion rule is based on the original data feature extraction layer, which can effectively solve the model collaboration problem of similar datasets in data heterogeneity.
Meanwhile, the generic fusion rule designed in (\ref{e11}) argues that the contributions of all models are equivalent. 
The idea is then applied to the fully connected layer, which means that the overfitting caused by the exclusion of non-similarity layers in personalized fusion rules can be mitigated.
                                                 
\section{Experimental Result}
In this section, three illustrative instances are given to demonstrate the superiority of the developed pFedCFR to the SOTA PFL methods.

\subsection{Experimental Setup}
Note that the software/hardware configuration of the system in the experiment is as follows.
The program is executed by using the framework of Pytorch 1.9, which runs on the server system of Ubuntu 20.04.3 LTS with 512G memory, NVIDIA 3080 GPU and Intel Core-i7 CPU@3.6GHz.

\subsubsection{Dataset Description}
Three public benchmark datasets were used in the experiments, they are MNIST \cite{a20}, FMNIST \cite{a21} and CIFAR-10 \cite{a22}, and the specific statistical properties are shown in TABLE~{\ref{table1}}.
Moreover, each dataset was preprocessed with normalization before segmentation and training.

\begin{table}
\caption{Dataset Characteristic}
\label{table1}
\centering
\begin{tabular}{ccccl}
\toprule
\bf{Dataset}  	&\bf{MNIST}		&\bf{FASHION-MNIST}		&\bf{CIFAR-10}\\
\midrule
Items       	&70000		&70000				&60000\\	
Class			&10			&10					&10		\\
Dimension       &(28,28)	&(28,28)			&(3,32,32)		\\
Train/Test      &(6:1)		&(6:1)			&(5:1)	\\
Intro			&numerical 	&clothes				&animals and vehicles			\\
\bottomrule
\end{tabular}
\end{table}

Owing to the limitation of computational resources, the dataset is divided according to the requirement of 20 clients in this paper.
Meanwhile, considering the non-independent and identical distribution of samples in practice, each client is allocated with only partially labelled training samples, and the sample capacity size of each client varies widely.
Specifically, by using the heterogeneous data construction rules in \cite{a12}, we first assign corresponding labels to each client, then divide the number of samples using the strategy of combining lognormal distribution and random factor, and finally achieve the segmentation of all client samples.

\subsubsection{Compared Methods and Hyperparameters}
To fully and comprehensively show the superiority of the proposed method, pFedCFR is experimentally compared with four mainstream methods in this paper. It should be pointed out that the hyperparameter settings in these methods are adopted from the original proposal.
And the specific details are as follows.
\begin{itemize}
\item [1)] FedAvg is one of the most common representations of federation learning \cite{a5}, and its fusion strategy is to average the model parameters uploaded by each client.
The learning rate $\eta$ is set to 0.005.
\item [2)] FedProx in \cite{fedprox} solves the data heterogeneity and convergence problem of client model updates by adding a global approximation penalty term.
Where the penalty coefficient $\mu=0.001$ and the learning rate $\eta=0.005$.
\item [3)] The goal of Ditto in \cite{a12} is to train the optimal private model for each client to meet the personalization needs of heterogeneous data.
Where the local step $l_{\lambda} = 1$ and the learning rate $\eta=0.005$.
\item [4)] pFedMe introduces the idea of personalization \cite{a13}, which transforms the optimization problem into a bi-level decoupling problem from client personalization loss to global loss.
Where the penalty coefficient $\lambda$ and the number of personalized training steps $K$ are set to $15$ and $5$, respectively.
In addition, the global tuning parameter $\beta=1$ and the personalized learning rate $\eta=0.005$.
\item [5)] FedAMP proposes a personalized fusion strategy with a model near-similarity-repelling-difference by introducing a message-passing mechanism \cite{a15}.
Since the method is sensitive to hyperparameters, the hyperparameter settings in the experiments were strictly adopted from the authors' suggestion, i.e. penalty coefficient $\lambda=1$, convergence coefficient $\alpha_k=10^4$, weighted hyperparameters $\sigma=10^6$, and the learning rate $\eta=0.005$.
\end{itemize}

Notice that the significant parameters of the proposed pFedCFR are configured in the following way.
According to the experimental validation and the convergence analysis in \cite{a15}, the convergence coefficient $\alpha_t$ and the personalized penalty coefficient $\lambda$ in (\ref{e10}) are set to $10^4$ and $1$, respectively, and the hyperparameter $\sigma$ in $A(x)$ is set to $10^6$, and the generic penalty term coefficient $\mu$ in (\ref{e12}) is set to $0.001$.
Moreover, the deep neural network (DNN) with 2-layer fully connected units is selected as the training network for datasets MNIST and Fashion-MNIST.
On the other hand, the convolutional neural network (CNN), which consists of 2 convolutional layers and 2-layer fully connected units, and the residual neural network (Resnet-18) in \cite{a23} are both selected as the model framework for the client in dataset CIFAR-10.
The client numbers $N=20$, the local update step is $10$, the global communication round $T=100$, and the learning rate $\eta=0.005$.
Finally, the core parameter personalized fusion threshold $r$ in Algorithm 1 is set to $1$ and $2$ in the DNN and CNN, respectively.

\subsection{Results on Heterogeneity Data}
The proposed pFedCFR is experimentally validated on several heterogeneous datasets, where the performance comparison results with various typical approaches are expressed in TABLE~{\ref{table2}}, and these results confirm that the proposed method is effective.
As can be seen from the table, pFedCFR outperforms the other fusion models in terms of prediction accuracy, especially in the CIFAR-10 (CNN) configuration setting, with an accuracy of 0.7809, which improves the accuracy value by 0.015 over the second-ranked FedAMP.
At the same time, it is noticed that FedAvg and FedProx with global optimization ideas in the experimental results are far inferior to pFedMe and FedAMP with personalized fusion strategy in terms of accuracy, which indicates that the PFL fusion strategy studied in this paper is urgent under the non-independent and identical distribution of heterogeneity data.
\begin{table*}
\caption{The results of prediction accuracy comparison of pFedCFR with several SOTA methods under different data sets and multiple network models. The bolded font indicates the best prediction performance of the method.}
\label{table2}
\centering
\begin{tabular}{ccccc}
\toprule
          &             & \multicolumn{3}{c}{{Dataset}} \\
                         \cline{3-5} 
&\multirow{-2}{*}{\bf{Method}} & \bf{MNIST (DNN)}  & \bf{Fashion-MNIST (DNN)}  & \bf{CIFAR-10 (CNN)} \\
\midrule

&FedAvg \cite{a5}				&  $0.8502 \pm 0.0008$     	& $ 0.7034\pm 0.0014$			& $0.6081 \pm 0.0027$		\\
\multirow{-2}{*}{\bf{Global}}	&FedProx \cite{fedprox}		&  $0.8180\pm 0.0002$     	& $ 0.7059\pm 0.0024$			& $0.6073 \pm 0.0019$		\\
\midrule
&Ditto \cite{a12}			&  $0.9530\pm 0.0060$      	& $0.9498 \pm 0.0082$			& $0.7308 \pm 0.0100$		\\
&pFedMe \cite{a13}			&  $0.9363\pm 0.0160$      	& $0.9333 \pm 0.0020$			& $0.7693 \pm 0.0010$		\\
&FedAMP \cite{a15}			&  $0.9541\pm 0.0004$      	& $ 0.9535\pm 0.0006$			& $0.7656 \pm 0.0004$		\\
\multirow{-4}{*}{\bf{Personalized}}	&\bf{pFedCFR}	& {$\bf{0.9563} \pm 0.0008$}	& $ \bf{0.9592}\pm 0.0003$	& $\bf{0.7809} \pm 0.0010$	\\
\bottomrule
\end{tabular}
\end{table*}
Furthermore, in the study of personalized fusion strategies, FedAMP based on the message-passing mechanism has better personalization services, which are reflected in the classification accuracy of all datasets.
It is worth noting that the personalized in FedAMP and the generalized fusion rules in FedProx are improved to each layer in pFedCFR, and the above comparison results provide solid evidence for the advancedness of these improvements.

To illustrate that pFedCFR improves the overfitting defect under the strong personality rule, this paper conducts three experiments with the loss and accuracy comparison in the same heterogeneous configuration.
\begin{figure*}[!htb]
\subfigbottomskip=0pt
\centering
\subfigure[]{
\includegraphics[width=0.46\linewidth,keepaspectratio=true]{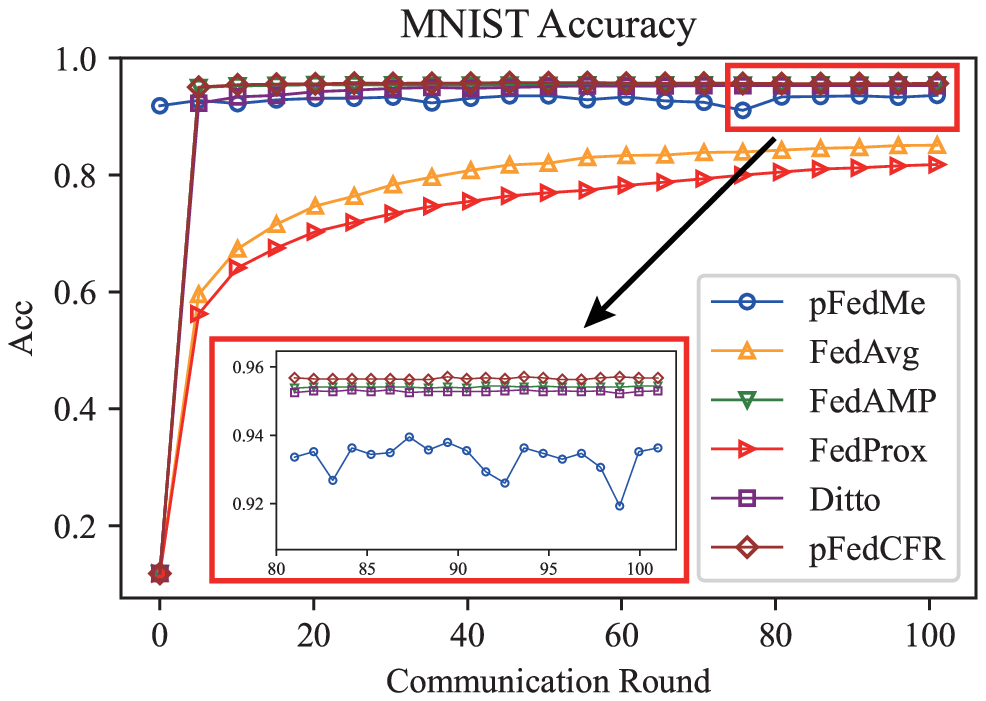}}
\subfigure[]{
\includegraphics[width=0.46\linewidth,keepaspectratio=true]{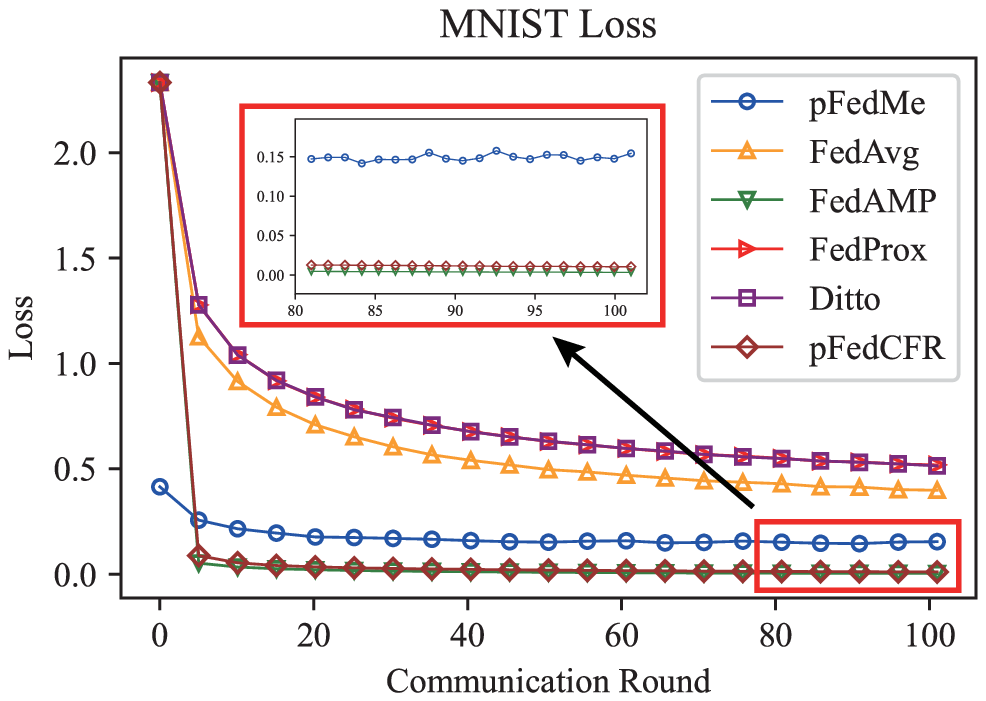}}
\subfigure[]{
\includegraphics[width=0.46\linewidth,keepaspectratio=true]{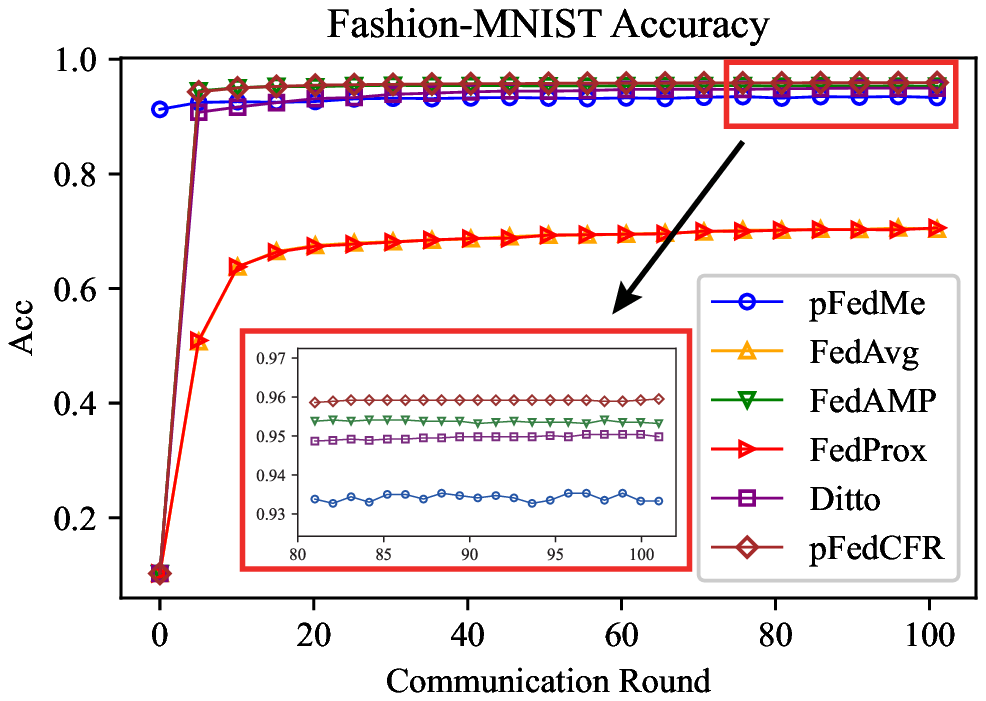}}
\subfigure[]{
\includegraphics[width=0.46\linewidth,keepaspectratio=true]{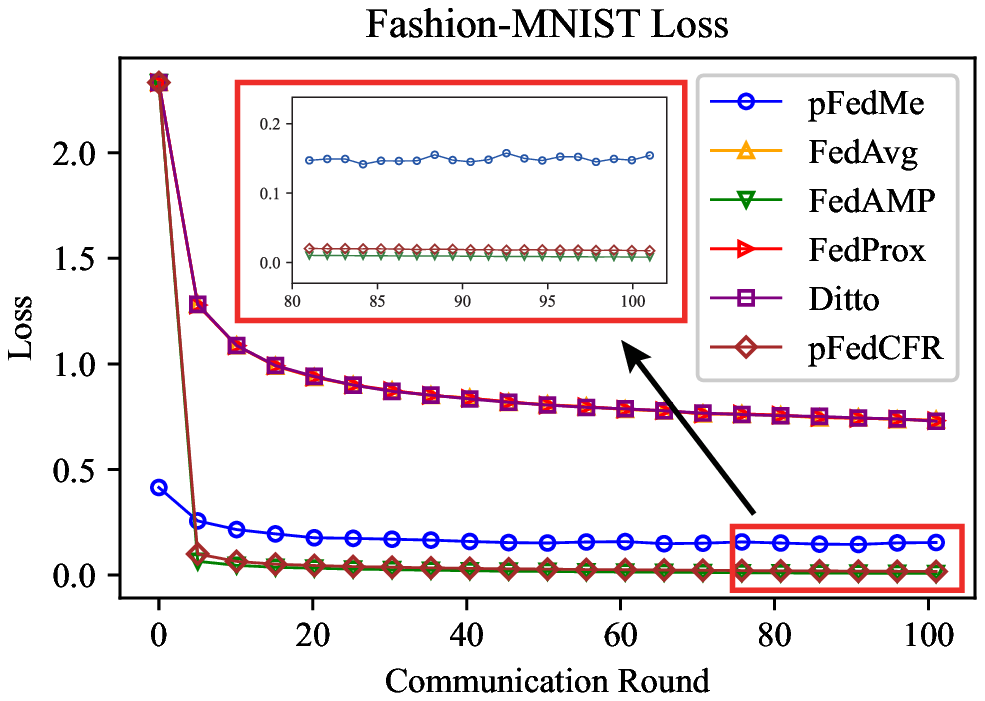}}
\subfigure[]{
\includegraphics[width=0.46\linewidth,keepaspectratio=true]{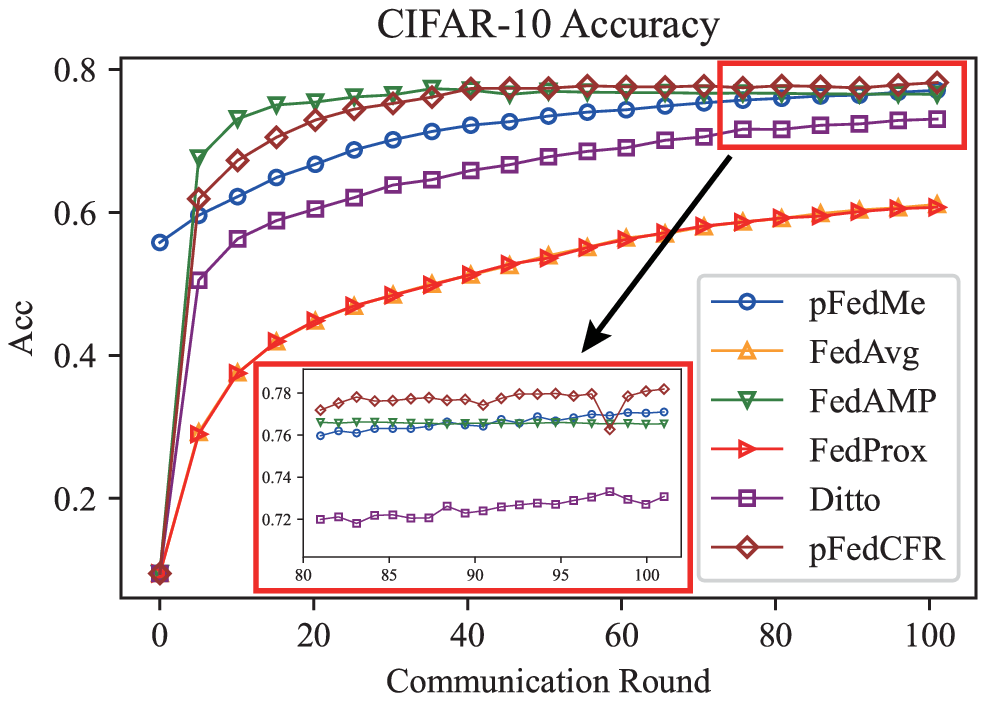}}
\subfigure[]{
\includegraphics[width=0.46\linewidth,keepaspectratio=true]{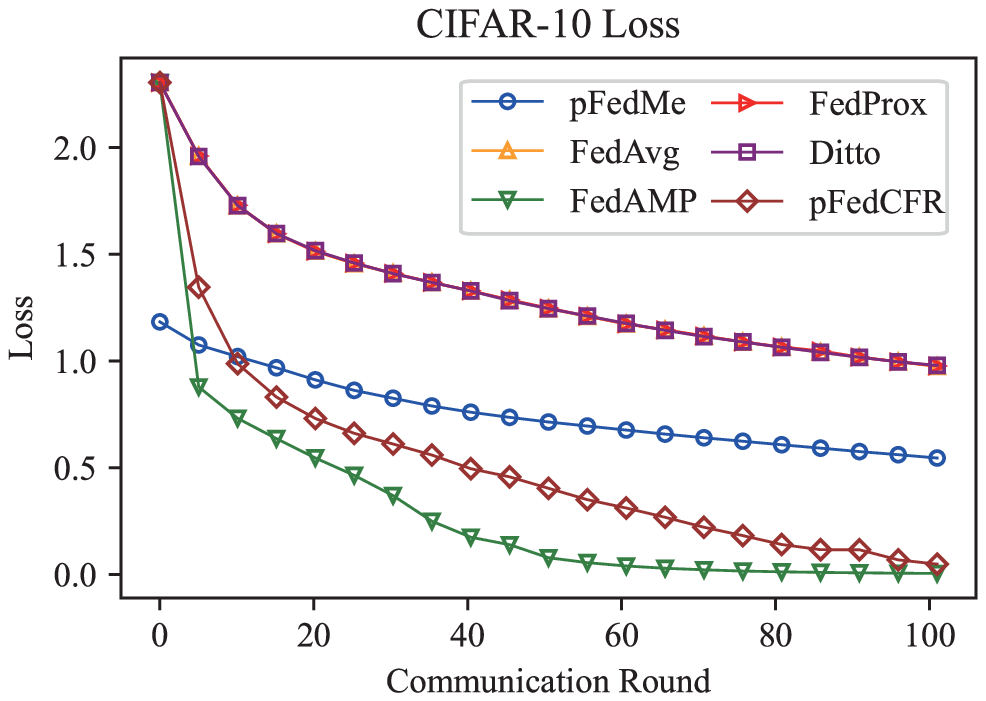}}
\caption{Comparison of test accuracy and loss between different federated learning methods within 100 communication rounds.}
\label{fig2}
\end{figure*}
As shown in Fig.~{\ref{fig2}}, the global federated learning methods suffer from underperformance as the number of communication rounds increases, while the PFL methods enter an overfitting state, where the red rectangular box is a detailed comparison of each method.  
In particular, the FedAMP consistently presents a decrease in loss without an increase in test accuracy after about 40 rounds.
This implied that while the message-passing strategy enhances feature collaboration between similar clients, it reduces model generality, which leads to overfitting.
Combining the test accuracy and loss comparison of the three benchmark experiments shows that the performance of Ditto needs to be improved, although no overfitting effect was observed.
Interestingly, it is shown from the experiments that the proposed pFedCFR has significant performance improvement and alleviates the overfitting phenomenon mentioned above.
Moreover, the test accuracy and stability of pFedCFR outperformed the pFedMe in all experiments.

\subsection{Cross-Fusion Operations in Different Layers}
According to the core idea of cross-fusion in (\ref{e14}), the proposed pFedCFR with layers as the basic fusion unit contains both personalized and generalized fusion strategies.
To show the effects of the fusion strategies employed at different network layers on the performance of the algorithms.
In this section, three experiments with Resnet-18 are compared to verify the effectiveness of the feature extraction layer using a personalized fusion strategy and the decision layer using a generic fusion strategy.

\begin{figure}[!htb]
\subfigbottomskip=0pt
\centering
\subfigure[]{
\includegraphics[width=1\linewidth,keepaspectratio=true]{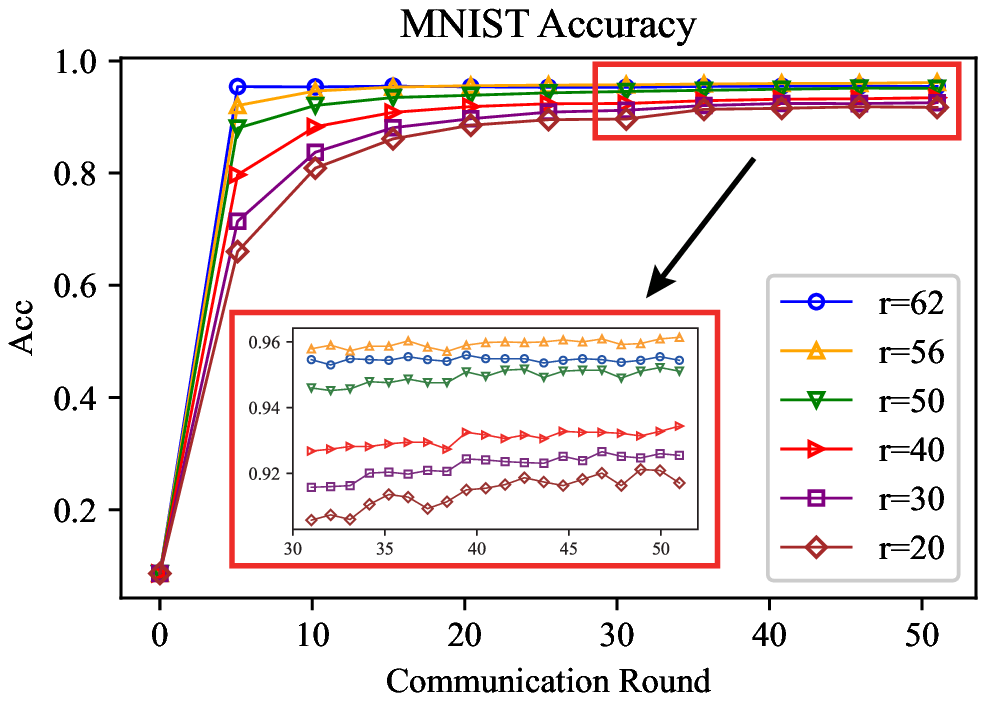}}
\subfigure[]{
\includegraphics[width=1\linewidth,keepaspectratio=true]{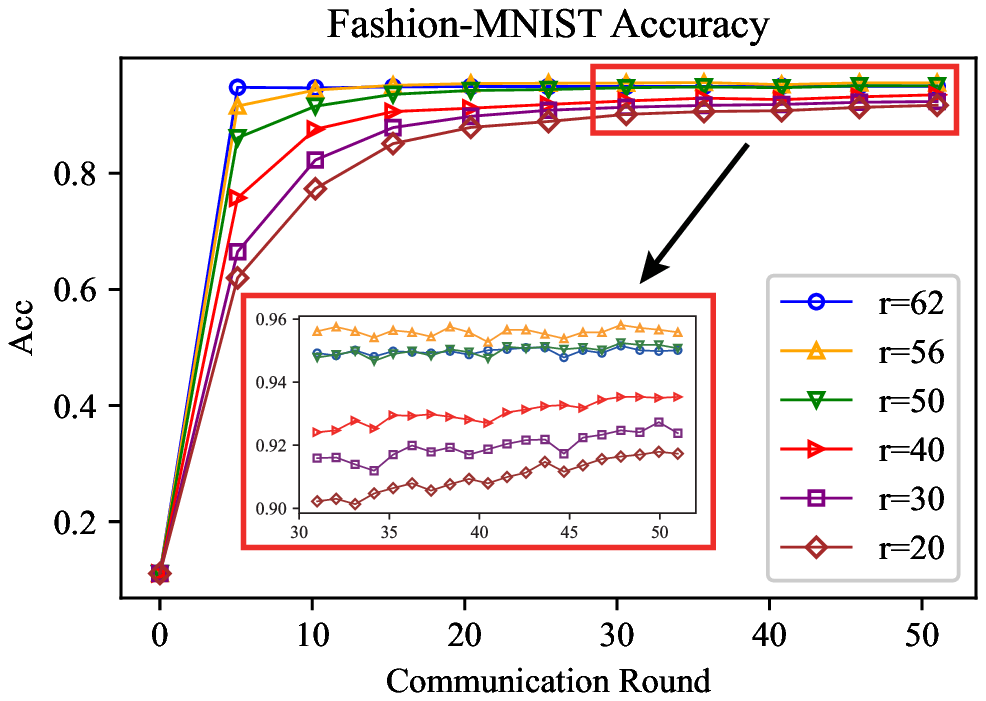}}
\subfigure[]{
\includegraphics[width=1\linewidth,keepaspectratio=true]{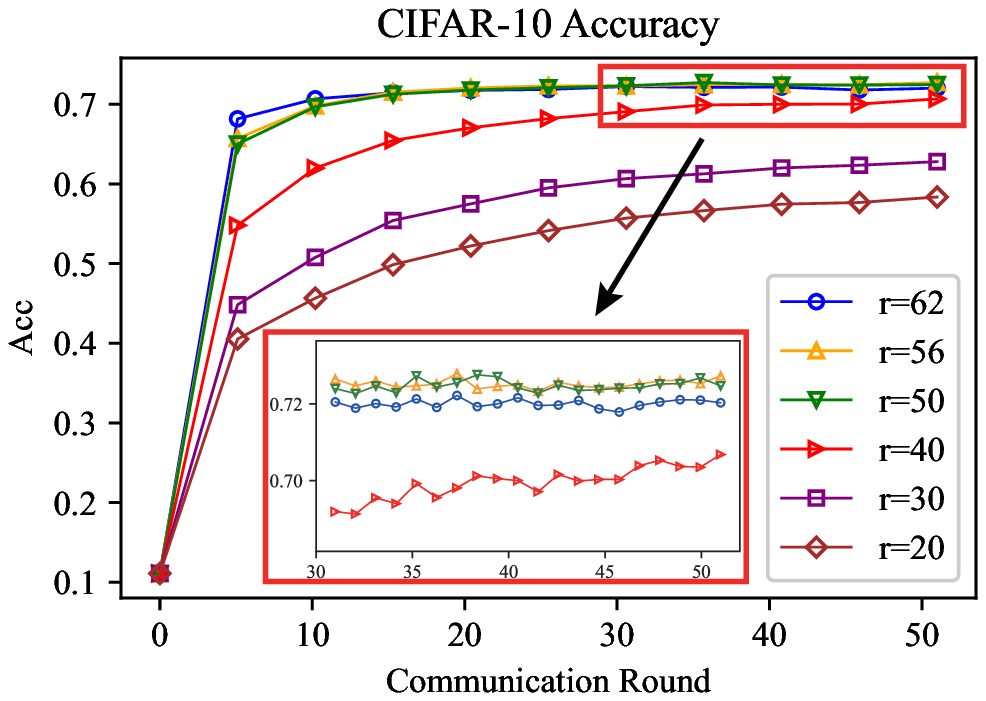}}\hspace{-0pt}
\caption{Accuracy comparison of the proposed pFedCFR method with different personalized fusion thresholds in three benchmark datasets, where the experimentally used Resnet-18 depth is 64.}
\label{fig3}
\end{figure}
\begin{table}
\caption{The test accuracy comparison of pFedCFR in deep learning model Resnet-18 with different personalized fusion threshold.}
\label{table3}
\centering
\begin{tabular}{ccccc}
\toprule
&{\bf{r}} & \bf{MNIST}  & \bf{Fashion-MNIST}  & \bf{CIFAR-10} \\
\midrule
&62			&  $0.9555\pm 0.0017$     	& $0.9515\pm 0.0017$			& $0.7211\pm 0.0015$		\\
&56			&  $\bf{0.9614}\pm 0.0004$     	& $\bf{0.9581}\pm 0.0033$			& $\bf{0.7273}\pm 0.0021$		\\
&50			&  $0.9522\pm 0.0032$      	& $0.9524\pm 0.0023$			& $0.7268\pm 0.0027$		\\
&40			&  $0.9344\pm 0.0019$      	& $0.9353\pm 0.0035$			& $0.7068\pm 0.0065$		\\
&30			&  $0.9266\pm 0.0014$      	& $0.9273\pm 0.0040$			& $0.6285\pm 0.0018$		\\
&20			&  $0.9212\pm 0.0039$		& $0.9179\pm 0.0023$			& $0.5836\pm 0.0045$	\\
\bottomrule
\end{tabular}
\end{table}
TABLE~{\ref{table3}} reports the test accuracy of pFedCFR as the personalized fusion threshold $r$ in the range $\left\{20, 30, 40, 50, 56, 62\right\}$, where the full model depth $L$ in Resnet-18 is $64$, and $l\leq r$ personalized fusion strategy is adopted, while $l>r$ the generic fusion strategy is utilized.
Since the first 56 layers of the model in Resnet-18 are convolutional feature extraction layers, while the remaining are full-connected decision calculation layers.
The result in TABLE~{\ref{table3}} illustrates that the best performance of pFedCFR is when $r=56$, which is consistent with the expectation of personalized fusion threshold selection in this paper.
It also indicates that as $r$ decreases, more personalized fusion network layers are replaced by generic fusion, which consequently causes a decrease in test accuracy.

The experimental results show that an accurate selection of personalized fusion thresholds is important for pFedCFR.
To depict the effect of different thresholds on the model training process in detail, information from 50 communication rounds was experimentally recorded, as shown in Fig.~{\ref{fig3}}.
Apparently, it clearly implies that during the training process for MNIST and Fashion-MNIST, the model performance is generally ahead of the other thresholds when $r$ is taken as $56$.
Although there was a crossover between $r=56$ and $r=50$ in the CIFAR-10 experiment, the former could be observed to be superior overall.
Combined with the analysis in Section IV-B, overfitting is the main reason for the performance degradation at $r = 62$.
With the basic model determined, these results support us in quickly determining the personalized fusion threshold of the pFedCFR.

\section{Conclutions}
In this paper, a new PFL method called pFedCFR has been developed for the data heterogeneity problem among multiple clients.
The designed multi-layer multi-fusion strategies framework based on layer functions effectively improves the low-performance problem caused by single fusion policy in existing federated learning.
Then, a fusion strategy combining personalization and generalization is designed to alleviate the overfitting phenomenon caused by forceful personalized mechanisms.
The extended experiments demonstrate the effectiveness of the proposed method.

On the other hand, consider the following critical issues in the PFL algorithm:
i). although the overfitting phenomenon is alleviated, it still exists in the forceful personalized fusion rules,
ii). only two functions, the feature extraction layer and the decision layer, are considered.
Therefore, we will work on a more detailed and generic fusion strategy based on pFedCFR.


\bibliographystyle{ieeetr} 

\bibliography{mybibfile}

\begin{thebibliography}{10}

\bibitem{a1}
Q.~Yang, Y.~Liu, T.~Chen, and Y.~Tong, ``Federated machine learning: Concept
  and applications,'' {\em ACM Trans. Intell. Syst. Technol.}, vol.~10, jan
  2019.

\bibitem{kbs1}
C.~Zhang, Y.~Xie, H.~Bai, B.~Yu, W.~Li, and Y.~Gao, ``A survey on federated
  learning,'' {\em Knowledge-Based Systems}, vol.~216, p.~106775, 2021.

\bibitem{a2}
M.~Hao, H.~Li, X.~Luo, G.~Xu, H.~Yang, and S.~Liu, ``Efficient and
  privacy-enhanced federated learning for industrial artificial intelligence,''
  {\em IEEE transactions on industrial informatics}, vol.~16, no.~10,
  pp.~6532--6542, 2020.

\bibitem{kbs2}
T.~Shaik, X.~Tao, N.~Higgins, R.~Gururajan, Y.~Li, X.~Zhou, and U.~R. Acharya,
  ``Fedstack: Personalized activity monitoring using stacked federated
  learning,'' {\em Knowledge-Based Systems}, vol.~257, p.~109929, 2022.

\bibitem{a4}
X.~Tu, K.~Zhu, N.~C. Luong, D.~Niyato, Y.~Zhang, and J.~Li, ``Incentive
  mechanisms for federated learning: From economic and game theoretic
  perspective,'' {\em IEEE transactions on cognitive communications and
  networking}, vol.~8, no.~3, pp.~1--1, 2022.

\bibitem{a28}
J.~D. Fernandez, S.~P. Menci, C.~M. Lee, A.~Rieger, and G.~Fridgen,
  ``Privacy-preserving federated learning for residential short-term load
  forecasting,'' {\em Applied energy}, vol.~326, p.~119915, 2022.

\bibitem{a29}
W.~Zhang, X.~Li, H.~Ma, Z.~Luo, and X.~Li, ``Federated learning for machinery
  fault diagnosis with dynamic validation and self-supervision,'' {\em
  Knowledge-Based Systems}, vol.~213, p.~106679, 2021.

\bibitem{a5}
H.~B. McMahan, E.~Moore, D.~Ramage, S.~Hampson, and B.~A. y~Arcas,
  ``Communication-efficient learning of deep networks from decentralized
  data,'' in {\em 2017 20th International Conference on Artificial Intelligence
  and Statistics}, pp.~1273--1282, 2017.

\bibitem{a6}
X.~Zhang, Y.~Li, W.~Li, K.~Guo, and Y.~Shao, ``Personalized federated learning
  via variational bayesian inference,'' in {\em 39th International Conference
  on Machine Learning (ICML)}, 2022.

\bibitem{a25}
F.~Sattler, S.~Wiedemann, K.-R. Müller, and W.~Samek, ``Robust and
  communication-efficient federated learning from non-i.i.d. data,'' {\em IEEE
  Transactions on Neural Networks and Learning Systems}, vol.~31, no.~9,
  pp.~3400--3413, 2020.

\bibitem{a26}
H.~Jamali-Rad, M.~Abdizadeh, and A.~Singh, ``Federated learning with taskonomy
  for non-iid data,'' {\em IEEE Transactions on Neural Networks and Learning
  Systems}, pp.~1--12, 2022.

\bibitem{kbs3}
Y.~Zhang, S.~Wei, S.~Liu, Y.~Wang, Y.~Xu, Y.~Li, and X.~Shang,
  ``Graph-regularized federated learning with shareable side information,''
  {\em Knowledge-Based Systems}, vol.~257, p.~109960, 2022.

\bibitem{a7}
Y.~Park and J.~C. Ho, ``Tackling overfitting in boosting for noisy healthcare
  data,'' {\em IEEE transactions on knowledge and data engineering}, vol.~33,
  no.~7, pp.~2995--3006, 2021.

\bibitem{fedprox}
T.~Li, A.~K. Sahu, M.~Zaheer, M.~Sanjabi, A.~Talwalkar, and V.~Smith,
  ``Federated optimization in heterogeneous networks,'' in {\em Proceedings of
  Machine Learning and Systems}, vol.~2, pp.~429--450, 2020.

\bibitem{a24}
A.~Z. Tan, H.~Yu, L.~Cui, and Q.~Yang, ``Towards personalized federated
  learning,'' {\em IEEE transaction on neural networks and learning systems},
  vol.~PP, pp.~1--17, 2022.

\bibitem{a8}
F.~Hanzely, S.~Hanzely, S.~Horv\'{a}th, and P.~Richtarik, ``Lower bounds and
  optimal algorithms for personalized federated learning,'' in {\em Advances in
  Neural Information Processing Systems}, vol.~33, pp.~2304--2315, 2020.

\bibitem{a9}
A.~Fallah, A.~Mokhtari, and A.~Ozdaglar, ``Personalized federated learning with
  theoretical guarantees: A model-agnostic meta-learning approach,'' in {\em
  Advances in Neural Information Processing Systems}, vol.~33, pp.~3557--3568,
  2020.

\bibitem{a10}
J.~Mills, J.~Hu, and G.~Min, ``Multi-task federated learning for personalised
  deep neural networks in edge computing,'' {\em IEEE transactions on parallel
  and distributed systems}, vol.~33, no.~3, pp.~630--641, 2022.

\bibitem{a12}
T.~Li, S.~Hu, A.~Beirami, and V.~Smith, ``Ditto: Fair and robust federated
  learning through personalization,'' in {\em International Conference on
  Machine Learning (ICML)}, 2021.

\bibitem{a13}
C.~T.~Dinh, N.~Tran, and J.~Nguyen, ``Personalized federated learning with
  moreau envelopes,'' in {\em Advances in Neural Information Processing
  Systems}, vol.~33, pp.~21394--21405, 2020.

\bibitem{a14}
Y.~Zhao, G.~Yu, J.~Wang, C.~Domeniconi, M.~Guo, X.~Zhang, and L.~Cui,
  ``Personalized federated few-shot learning,'' {\em IEEE transaction on neural
  networks and learning systems}, vol.~PP, pp.~1--11, 2022.

\bibitem{a15}
Y.~Huang, L.~Chu, Z.~Zhou, L.~Wang, J.~Liu, J.~Pei, and Y.~Zhang,
  ``Personalized cross-silo federated learning on non-iid data,'' {\em
  Proceedings of the AAAI Conference on Artificial Intelligence}, vol.~35,
  pp.~7865--7873, May 2021.

\bibitem{pfedfomo}
M.~Zhang, K.~Sapra, S.~Fidler, S.~Yeung, and J.~M. Alvarez, ``Personalized
  federated learning with first order model optimization,'' in {\em
  International Conference on Learning Representations}, 2021.

\bibitem{a16}
I.~Achituve, A.~Shamsian, A.~Navon, G.~Chechik, and E.~Fetaya, ``Personalized
  federated learning with gaussian processes,'' in {\em Advances in Neural
  Information Processing Systems}, vol.~34, pp.~8392--8406, 2021.

\bibitem{a17}
R.~Wu, A.~Scaglione, H.-T. Wai, N.~Karakoc, K.~Hreinsson, and W.-K. Ma,
  ``Federated block coordinate descent scheme for learning global and
  personalized models,'' {\em Proceedings of the AAAI Conference on Artificial
  Intelligence}, vol.~35(12), pp.~10355--10362, 2021.

\bibitem{a18}
J.~Sun, Y.~Li, H.~Chen, B.~Zhang, and J.~Zhu, ``Memf: Multi-level-attention
  embedding and multi-layer-feature fusion model for person
  re-identification,'' {\em Pattern recognition}, vol.~116, p.~107937, 2021.

\bibitem{a19}
Q.~Guo, S.~Qi, S.~Qi, D.~Wu, and Q.~Li, ``Fedmcsa: Personalized federated
  learning via model components self-attention,'' in {\em arXiv}, 2022.

\bibitem{a27}
Y.~Chen, X.~Sun, and Y.~Jin, ``Communication-efficient federated deep learning
  with layerwise asynchronous model update and temporally weighted
  aggregation,'' {\em IEEE Transactions on Neural Networks and Learning
  Systems}, vol.~31, no.~10, pp.~4229--4238, 2020.

\bibitem{a20}
Y.~Lecun, L.~Bottou, Y.~Bengio, and P.~Haffner, ``Gradient-based learning
  applied to document recognition,'' {\em Proceedings of the IEEE}, vol.~86,
  no.~11, pp.~2278--2324, 1998.

\bibitem{a21}
H.~Xiao, K.~Rasul, and R.~Vollgraf, ``Fashion-mnist: a novel image dataset for
  benchmarking machine learning algorithms,'' 2017.

\bibitem{a22}
A.~Krizhevsky, ``Learning multiple layers of features from tiny images,'' 2009.

\bibitem{a23}
K.~He, X.~Zhang, S.~Ren, and J.~Sun, ``Deep residual learning for image
  recognition,'' in {\em 2016 IEEE Conference on Computer Vision and Pattern
  Recognition (CVPR)}, pp.~770--778, 2016.

\end{thebibliography}
\end{document}